\documentclass[letterpaper, 10 pt, conference]{ieeeconf}  

\usepackage{amsthm}
\usepackage{amsmath,amsfonts}
\usepackage{amssymb}
\usepackage{amsfonts}
\usepackage{textcomp}
\usepackage{stfloats}
\usepackage{url}
\usepackage{verbatim}
\usepackage{graphicx}
\usepackage{lineno}
\usepackage{xcolor, soul}
\usepackage{array}
\usepackage{booktabs}
\usepackage{makecell}
\usepackage{enumerate}
\usepackage{algorithm}
\usepackage{arydshln}
\usepackage{booktabs}
\usepackage{multirow}
\usepackage{algpseudocode}
\usepackage{amsthm}
\usepackage{graphicx}
\usepackage{epsfig}
\usepackage{cite}
\usepackage{tensor}
\usepackage{color}
\usepackage{float}
\usepackage{comment}
\usepackage{mathrsfs}
\usepackage{blindtext}
\usepackage{soul}
\usepackage{graphics} 
\usepackage{epsfig} 
\usepackage{mathptmx} 
\usepackage{times} 
\usepackage{amsmath}
\usepackage{amssymb}  
\usepackage{balance}
\soulregister{\cite}7 
\soulregister{\citep}7 
\soulregister{\citet}7 
\soulregister{\ref}7 
\soulregister{\pageref}7 
\graphicspath{{figs1/}}
\IEEEoverridecommandlockouts                             
\theoremstyle{definition}
\newtheorem{remark}{Remark}

\newcommand\T{{\hspace{-0pt}\intercal}}

\DeclareMathAlphabet{\mathcal}{OMS}{cmsy}{m}{n}

\algrenewcommand\algorithmicrequire{\textbf{Input:}}
\algrenewcommand\algorithmicensure{\textbf{Output:}}
\usepackage{hyperref}
\hypersetup{
hypertex=true, 
colorlinks=true, 
linkcolor=blue, 
anchorcolor=blue,
citecolor=blue
}

\title{\LARGE \bf
Revolutionizing Packaging: A Robotic Bagging Pipeline with
 Constraint-aware Structure-of-Interest Planning
}

\author{
Jiaming Qi$^{1}$, 
Peng Zhou$^{1}$,
Pai Zheng$^{2}$,
Hongmin Wu$^{3}$,
Chenguang Yang$^{4}$,
David Navarro-Alarcon$^{2}$,
and Jia Pan$^{1}$
\thanks{
This work is supported by the Innovation and Technology Commission of the HKSAR Government under the InnoHK initiative. \emph{(Corresponding author: Jia Pan.)}}
\thanks{$^{1}$The University of Hong Kong, Hong Kong. e-mail: jpan@hku.hk}
\thanks{$^{2}$The Hong Kong Polytechnic University, Hong Kong.}
\thanks{$^{3}$Guangdong Academy of Sciences, China.}
\thanks{$^{4}$University of Liverpool, United Kingdom.}
}

\begin{document}

\maketitle

\begin{abstract}
Bagging operations, common in packaging and assisted living applications, are challenging due to a bag's complex deformable properties. To address this, we develop a robotic system for automated bagging tasks using an adaptive structure-of-interest (SOI) manipulation approach.
Our method relies on real-time visual feedback to dynamically adjust manipulation without requiring prior knowledge of bag materials or dynamics. 
We present a robust pipeline featuring state estimation for SOIs using Gaussian Mixture Models (GMM), SOI generation via optimization-based bagging techniques, SOI motion planning with Constrained Bidirectional Rapidly-exploring Random Trees (CBiRRT), and dual-arm manipulation coordinated by Model Predictive Control (MPC).
Experiments demonstrate the system's ability to achieve precise, stable bagging of various objects using adaptive coordination of the manipulators. The proposed framework advances the capability of dual-arm robots to perform more sophisticated automation of common tasks involving interactions with deformable objects.

\end{abstract}

\section{Introduction}
\label{section1}

The field of deformable object manipulation (DOM) has garnered considerable attention for its potential to automate many advanced tasks in human environments. 
Everyday objects, from garments to soft furnishings, present highly deformable behaviors that complicate its automatic handling. 
Providing robots with the sufficient dexterity to manipulate this type of objects is crucial for their seamless integration into daily human environments.
However, due to their infinite degrees-of-freedom and nonlinear dynamics, most research work has focused on simpler cases, such as 1-D and 2-D deformable bodies. 
The manipulation of complex 3D deformable structures such as common household bags (whose topology is modelled as a 2-torus), remains an underexplored problem in the robotics research community.

To address this gap in the literature, our work introduces a dual-arm robotic system empowered by constraint-aware structure-of-interest (SOI) planning, which advances DOM into the realm of 3D objects. This system is a significant leap towards sophisticated automation, capable of performing intricate tasks such as robotic bagging with precision and adaptability, marking a pivotal step in DOM research. In this paper, we introduce a novel approach to this problem through a dual-arm robotic system that leverages constraint-aware structure-of-interest (SOI) planning.

\begin{figure}[htbp]
\centering
\includegraphics[width=0.94 \columnwidth]{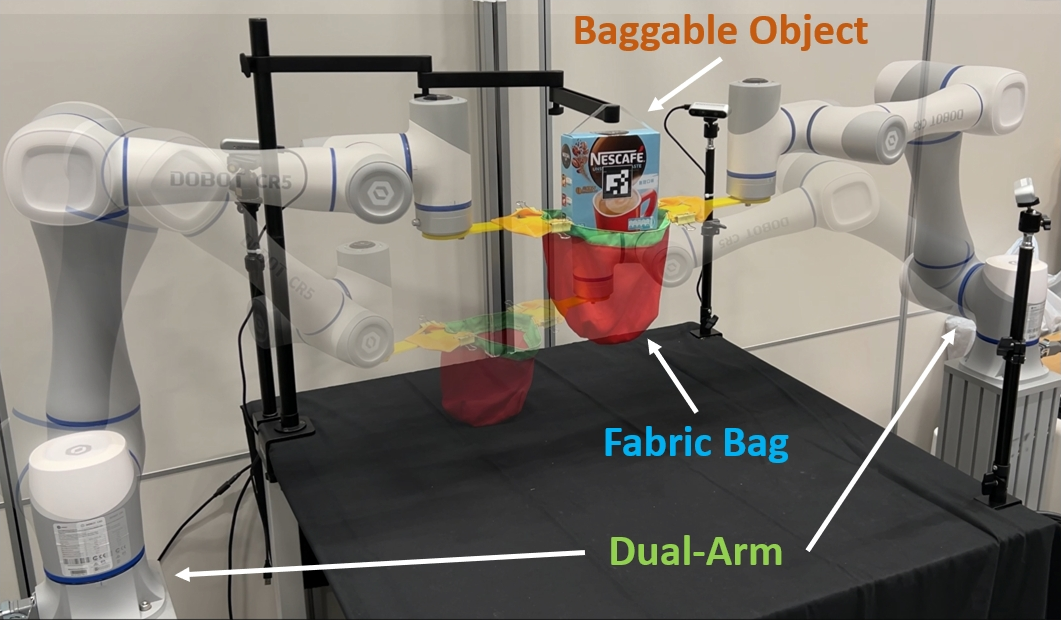}
\vspace{-0.3cm}
\caption{
The dual-arm grasps two handles of a fabric bag to manipulate the SOI (i.e., the opening rim) for the bagging task.
}
\vspace{-0.2cm}
\label{fig_example}
\end{figure}

Our approach is based on the insight that similar to the concept of Region of Interest (ROI) in the image processing domain, complete state estimation of a manipulated deformable object is not essential for robotic interaction. For specific deformable object manipulation tasks, it is sufficient to focus exclusively on state estimation of the critical structure-related components.
Take, for instance, a robotic bagging task: the opening rim of the fabric bag can be considered the Structure of Interest (SOI). By concentrating on state estimation related to just the opening rim, the robotic system can successfully accomplish the bagging task. This targeted approach simplifies the state estimation process and improves the efficiency and effectiveness of the manipulation task.
This system is specifically designed to address the automation of bagging tasks, a common yet challenging operation in both industrial and everyday contexts. The core of our approach is the use of two robotic arms that work in unison, guided by a sophisticated planning system that accounts for the constraints imposed by the object's structure and desired final state. This is achieved using 3D-printed connectors, which allow the robots to manipulate the bag with an unprecedented level of precision and stability.

Our contribution is as follows:
\begin{itemize}
    \item We propose a constraint-aware SOI planning framework that enables dual-arm robots to perform complex bagging tasks by manipulating a bag over an object to achieve a desired configuration.
    \item We integrate an adaptive vision-based control system that does not require prior knowledge of the bag’s material properties or system dynamics, making the setup more flexible and broadly applicable.
    \item We present a comprehensive methodological framework that encompasses SOI state estimation, bagging SOI generation, SOI planning, and motion planning, evidencing the system's adaptability and sensitivity to environmental constraints.
\end{itemize}

\section{Related Work}
\label{section2}
The manipulation of deformable objects by robotic systems has been an area of increasing interest within the robotics community \cite{gonnochenko2021coinbot}.
Early research efforts primarily addressed the manipulation of 1-D and 2-D deformable objects \cite{saha2007manipulation}, using techniques such as tension-based strategies \cite{kudo2000multi} and computational geometry \cite{alami1990geometrical} to model and control the behaviour of ropes, cloths, and sheets \cite{nair2017combining}.

With the shift towards 3D deformable objects, researchers have explored various methods to handle increased complexity \cite{seita2021learning}. 
Notably, work by \cite{wijayarathne2023real} delved into the dynamics of soft body manipulation using dual-arm robots, while \cite{zhang2023visual} focused on non-prehensile manipulation techniques for cloth folding tasks \cite{weng2024interactive}.
Both approaches laid the groundwork for understanding the intricate interplay between robotic control and deformable object dynamics.

Recent advancements in vision-based control systems, such as those by \cite{chen2023autobag}, have shown how real-time feedback can enhance the adaptability of robots to the unpredictable nature of deformable objects \cite{bahety2023bag,gu2024shakingbot}.
These works have informed the development of our constraint-aware SOI planning, which integrates real-time visual servoing to adjust the robot's actions on the fly.

Our work builds on these foundational studies and takes a significant step forward by focusing on dual-arm manipulation for the specific task of robotic bagging—a complex application that has received limited attention thus far. We leverage the principles of constraint-aware planning to address the intricate problem of enveloping 3D objects with a deformable bag, which requires a high level of coordination and sensitivity to the dynamic constraints of the object and its environment.

\vspace{-0.4cm}
\begin{figure}[htbp]
\centering
\includegraphics[width=0.85\columnwidth]{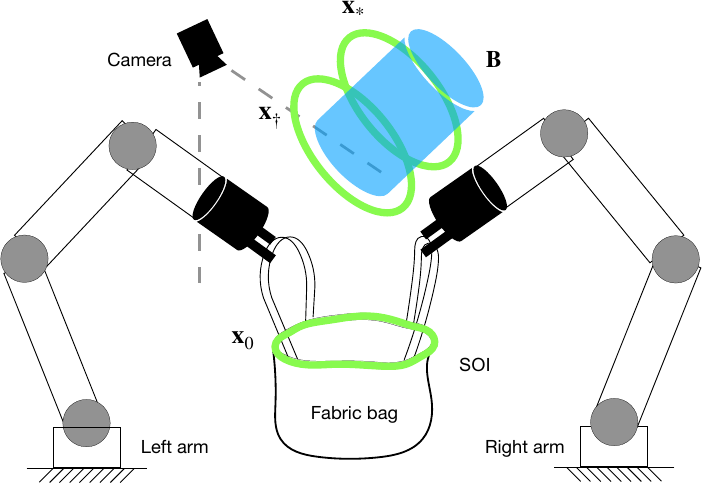}
\vspace{-0.2cm}
\caption{
The dual robot grasp two handles of a deformable fabric bag to manipulate the SOI (i.e., the opening rim) for the bagging task. 
}
\vspace{-0.3cm}
\label{fig1:bag_problem}
\end{figure}

\vspace{-0.3cm}
\section{Problem Statement}
\label{section3}

\emph{Notation.}
Subscript $(\cdot)_t$ is the discrete-time instant.
$\mathbf{I}_{n \times m}$ is the $n\times m$ matrix of ones, and the identity matrix as $\mathbf{E}_n$.
$\mathbf{L}_n$ is the low triangle matrix of $\mathbf{I}_{n \times n}$, and $\otimes$ is the Kronecker product.
$ ^{\mathcal{F}_x} \mathbf{p}$ is a point $\mathbf{p}$ in the frame $\mathcal{F}_x$.
In this work, unless otherwise specified, all points are expressed in the world frame $\mathcal{F}_w$ and omitted for clear.

In this work, we consider a novel robotic bagging task, as illustrated in Fig.~\ref{fig1:bag_problem}.
We propose a dual-arm robotic system where two robot manipulators grasp a deformable fabric bag to envelop a baggable object, denoted by $\mathbf{B}$, which is suspended in the air.
The system's objective is to manipulate the fabric bag from an initial state to a goal state, where the bag completely covers the object.
We posit that it is unnecessary to estimate the entire fabric bag; instead, focusing on the Section of Interest (SOI) of the bag, specifically the bag opening rim, is sufficient for this task.
Consequently, we define the SOI state of the bagging task as the opening rim of the bag, represented as $\mathbf{x}$, which can be captured by a depth camera configured in an eye-to-hand calibration style.
In contrast to \cite{zhou2024bimanual}, which uses the entire point cloud as the SOI, we opt for a simpler representation by selecting contour keypoints to depict our SOI:
\begin{equation}
\label{eq1}
\mathbf{x} = \left[ \mathbf{x}_1^\T, \ldots, \mathbf{x}_{n_x}^\T \right]^\T
\in \mathbb{R}^{3n_x},\quad
\mathbf{x}_i = \left[ x_i, y_i, z_i \right]^\T
\in \mathbb{R}^3
\end{equation}
where $n_x$ denotes the number of contour keypoints, and $\mathbf{x}_i$ represents the Cartesian coordinates of the $i$-th point in $\mathcal{F}_w$.

The task's goal is to manipulate the SOI of the bag from its initial state $\mathbf{x}_0$ to the target state $\mathbf{x}^*$.
We address this problem by planning with constraints:
1) The SOI keypoints should approximate the shape of an oval;
2) The perimeter formed by the SOI keypoints must be constant, indicating that the size of the fabric bag's opening rim does not change during manipulation.
Depending on whether the bag is in contact with the object $\mathbf{B}$, we divide the planning into \textit{pre-bagging} and \textit{bagging} stages:
\begin{equation}
\mathcal{G} := \left[ \mathcal{G}_{\text{pre-bagging}}, \mathcal{G}_{\text{bagging}} \right] \Big| \underbrace{\mathbf{g}_0, \mathbf{g}_1, \ldots, \mathbf{g}^{\dag}}_{\text{pre-bagging}}, \underbrace{\mathbf{g}^{\dag}, \ldots, \mathbf{g}^{\ast}}_{\text{bagging}},
\end{equation}
where $\mathbf{g}^{\dag}$ is an SOI shape tailored to the baggable object $\mathbf{B}$ that can perfectly envelop its bottom.
To reach each subgoal $\mathbf{g}_i$, we employ an MPC-based shape servoing approach to generate the velocity command $\mathbf{u}$ based on a measurable error function $\mathcal{E}_{\text{subgoal}}(\cdot)$ between the resulting SOI state $\mathbf{x}_i$ and the current subgoal $\mathbf{g}_i$:
\begin{equation}
\mathbf{u}_i = \underset{\mathbf{u} \in \mathcal{A}}{\arg\min}~\mathcal{E}_{\text{subgoal}}(\mathbf{x}_i, \mathbf{g}_i),
\label{eq:velocity_command}
\end{equation}
The bagging task is thus accomplished through a sequence of actions $\{\mathbf{u}_1, \mathbf{u}_2, \ldots, \mathbf{u}^*\}$.




\begin{figure*}[htbp]
\centering
\includegraphics[width=0.99\textwidth]{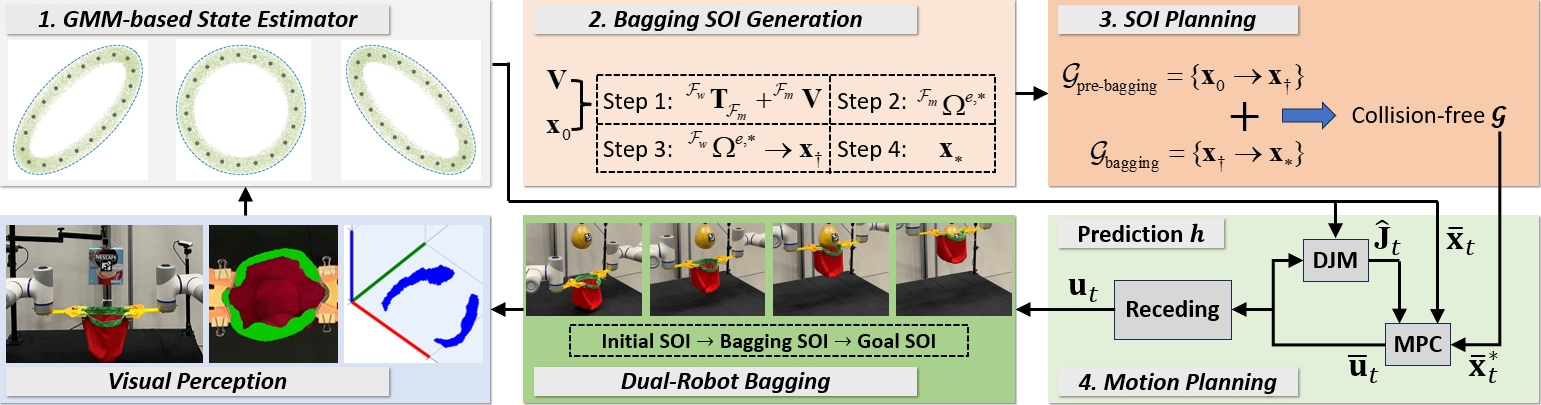}
\caption{
Schematic diagram of dual-arm manipulation approach of the bagging task.
The algorithm aims to command the robot to manipulate the bag into the specified shape to cover the bottom part of $\mathbf{B}$, i.e., bagging manner.
Experiments are conducted in the Cartesian space.
}
\vspace{-0.3cm}
\label{fig1}
\end{figure*}

\vspace{-0.1cm}
\section{Methodology}\label{section4}

In this section, a dual-arm manipulation approach of the bagging task is proposed, which is comprised of:
1) \textbf{SOI State Estimation.}
Extracting meaningful representations of the Structure of Interest
(SOI) from the raw dense and noisy point cloud. 
2) \textbf{Bagging SOI Generation.}
Generate a pre-enclosing shape $\mathbf{x}_{\ast}$ to cover the bottom part of  $\mathbf{B}$.
3) \textbf{SOI Planning.}
Generate a collision-free deformation path $\mathcal{G}$ from $\mathbf{x}_0$ to $\mathbf{x}_{\ast}$.
4) \textbf{Motion Planning.}
Formulating the bagging process as shape servoing, drive the dual-arm move along $\mathcal{G}$ to complete the bagging task.
Fig. \ref{fig1} presents the block diagram of the proposed manipulation approach.

\vspace{-0.3cm}
\subsection{SOI State Estimation}
\label{section4a}
In this work, the bag's SOI state is defined as a sequence of contour keypoints, i.e., $\mathcal{Q}_t = \{ \mathbf{x}_t^i \}, i \in [1, n_x]$.
The raw point cloud perceived by the depth camera is 
$\mathcal{P}_t = \{\mathbf{p}_t^i \}, i  \in [1, n_p]$, usually $n_p \gg n_x$.
The state estimator aims to obtain a concise representation $\mathbf{x}_t^i$ by aligning $\mathcal{Q}_t$ to $\mathcal{P}_t$ in real-time.

We adopt the sampling approach in \cite{tang2022track}, i.e., Structure preserved registration (SPR), formulating the alignment process as a probability density estimation problem for Gaussian mixture model (GMM).
By treating $\mathcal{P}_t$ as the points randomly sampled from GMM, thereby obtaining Gaussian's centroids as $\mathcal{Q}_t$.
Considering that $\mathcal{P}_t$ is dense, noisy and contains outliers, a uniform distribution for $\mathcal{Q}_t$ is added to GMM.
The sampling probability of $\mathbf{p}^m_t$ is taken as below:
\begin{equation}
\label{eq48}
\kappa (\mathbf{p}^m_t) = \Sigma_{n=1}^{n_x+1} \kappa (n)   \kappa (\mathbf{p}^m_t | n)
\end{equation}
where $\kappa(n)$ is the sampling weight of $n$-th mixture component, and $\kappa (\mathbf{p}^m_t | n)$ denotes the sampling probability of  $\mathbf{p}^m_t$ from $n$-th mixture component.
Both are given in \cite{tang2022track}.

The optimal estimation of $\mathcal{Q}_t$ can be obtained by maximizing the log-likelihood function $\mathcal{O}$ of the observation process:
\begin{equation}
\label{eq10}
\mathcal{O}(\mathbf{x}_t^n) 
= 
\sum_{m=1}^{n_p}
\sum_{n=1}^ {n_x+1}
\kappa (n|\mathbf{p} _t^m)  \log ( \kappa (n) \kappa (\mathbf{p}_t^m | n))
\end{equation}

\begin{figure}[ht]
\centering
\includegraphics[width=0.97\columnwidth]{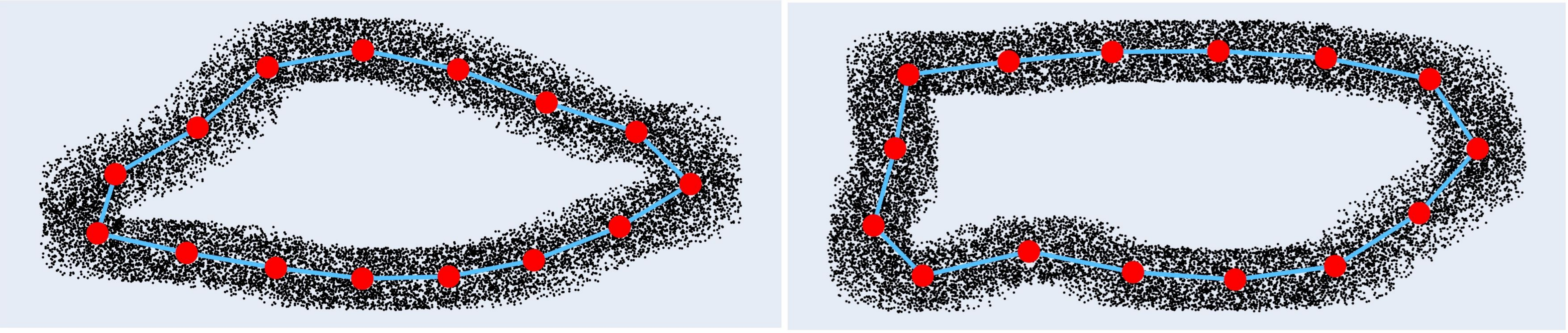}
\vspace{-0.2cm}
\caption{
The visual description of the GMM-based representation.
}
\vspace{-0.3cm}
\label{fig2}
\end{figure}

The maximization of \eqref{eq10} can be processed through the EM algorithm \cite{tang2022track}, the optimal result $\mathbf{x}_t^{n,\ast}$ is regarded the concise SOI representation of the bag.
Fig. \ref{fig2} visualizes the GMM-based representation, where the black dots are $\mathcal{P}_t$, and the red dots connected with the blue line are the precise $\mathcal{Q}_t$.

\begin{figure}[ht]
\centering
\includegraphics[width= 0.93\columnwidth]{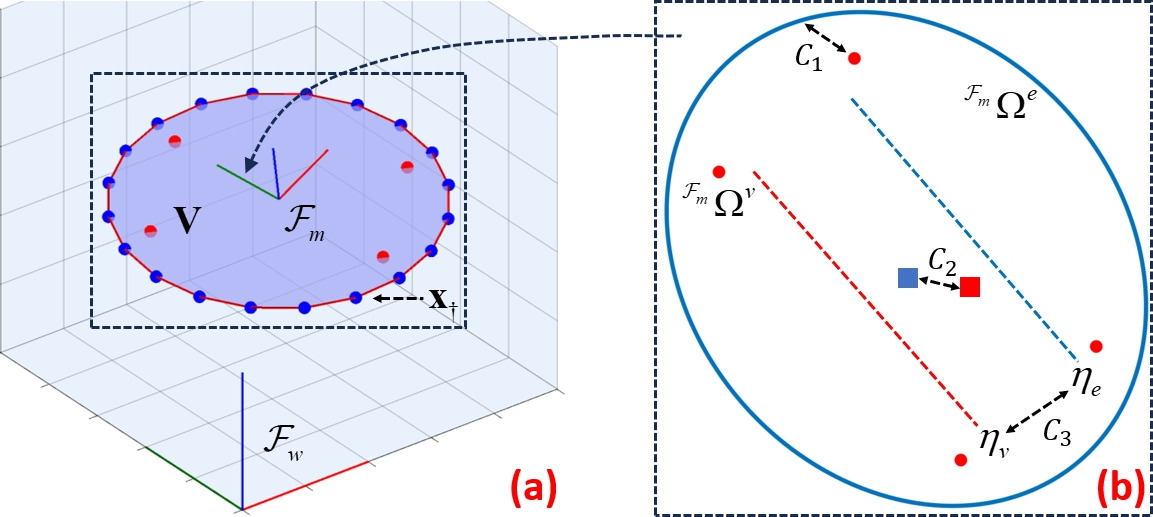}
\caption{
(a) The visualization of the bagging SOI.
(b) The projection of $\mathcal{F}_m$ in $\mathcal{F}_w$, and visualizing three constraints.
The red square is the centroid of $^{\mathcal{F}_m} \Omega^v$ and blue is that of $^{\mathcal{F}_m} \Omega^e$.
}
\vspace{-0.6cm}
\label{fig5}
\end{figure}

\vspace{-0.3cm}
\subsection{Bagging SOI Generation}\label{section4b}
This section introduces how to generate two shapes, i.e.,
a bagging SOI $\mathbf{x}_{\dag}$ covering the bottom of $\mathbf{B}$, while another is the goal SOI $\mathbf{x}_{\ast}$ for surrounding the entire lower part of $\mathbf{B}$.
As the dual robot manipulates the bag in an almost symmetrical way, thus the elliptical configuration is used as a reference to determine $\mathbf{x}_{\dag}$ and $\mathbf{x}_{\ast}$.
In this work, the bottom vertex points $\mathbf{V}$ of $\mathbf{B}$ are assumed to be coplanar, i.e., $\mathbf{V} = [\mathbf{v}_1, \ldots, \mathbf{v}_{n_v}]$, and $\mathbf{v}_i \in \mathbb{R}^3$ is the $i$-th vertex point's Cartesian coordinate in $\mathcal{F}_w$.
The essence is to change the SOI generation in $\mathcal{F}_w$ into 2D generation in the $xy$-plane in the mapping frame $\mathcal{F}_m$, which is built on the plane where $\mathbf{V}$ is located.

\textbf{Step 1:} 
\emph{Calculate $\mathcal{F}_m$ on the plane consisting  $\mathbf{V}$.}

Two auxiliary vectors are given as: 
${\xi}_i = \mathbf{v}_i - \bar{\mathbf{v}} , i=1,2.$ 
where $\bar{\mathbf{v}}$ is the centroid of $\mathbf{V}$.
The $z$-axis of $\mathcal{F}_m$ is calculated as ${\mathbf{a}} = ({\xi}_1 \times {\xi}_2)  /  \| {\xi}_1 \times {\xi}_2 \|$.
Any point is selected to determine the $y$-axis: ${\mathbf{o}} = (\mathbf{v}_3 - \bar{\mathbf{v}} ) /   \| \mathbf{v}_3 - \bar{\mathbf{v}} \|$.
Then, $x$-axis is given as
${\mathbf{n}} = ({\mathbf{o}} \times {\mathbf{a}} )   /   \| {\mathbf{o}} \times {\mathbf{a}} \|$.
For the uniqueness of $\mathcal{F}_m$, taking ${\mathbf{p}} = \bar{\mathbf{v}}$ as the originate of $\mathcal{F}_m$.
The transformation matrix from $\mathcal{F}_w$ to $\mathcal{F}_m$ is constructed as:
\begin{equation}
\label{eq16}
^{\mathcal{F}_w} \mathbf{T}_{\mathcal{F}_m} 
= 
\left[
\begin{array}{c:c:c:c}
{\mathbf{n}} & {\mathbf{o}} & {\mathbf{a}} & {\mathbf{p}}  \\
\hdashline
0 & 0 & 0 & 1 
\end{array}
\right] 
\in \mathbb{R}^{4 \times 4}
\end{equation}

Adopting \eqref{eq16} to map $\mathbf{V}$ into $\mathcal{F}_m$, denoted as $^{\mathcal{F}_m} \mathbf{V} = \{  ^{\mathcal{F}_m}  \mathbf{v}_i  \} \in \mathbb{R}^{n_v \times 3}$ where $^{\mathcal{F}_m}  \mathbf{v}_i =[^{\mathcal{F}_m} v_{i,x}, 
^{\mathcal{F}_m}  v_{i,y}, 
^{\mathcal{F}_m} v_{i,z}]$, then the centroid of $^{\mathcal{F}_m} \mathbf{V}$ is denoted as $^{\mathcal{F}_m} \bar{\mathbf{v}}$.
Normally, the $z$-axis of $^{\mathcal{F}_m} \mathbf{V}$ is close to zero, as $\mathbf{V}$ is assumed to be coplanar.



\textbf{Step 2:}
\emph{Calculate bagging ellipse in $xy$-plane of $\mathcal{F}_m$.}

The 2D ellipse parametric equation is constructed as:
\begin{align}
\label{eq18}
x 
&= \tau_x + \rho_a  \cos(\theta)  \cos(\alpha) - \rho_b \sin(\theta) \sin(\alpha) \notag
\\
y 
&= \tau_y + \rho_a  \cos(\theta) \sin(\alpha) + \rho_b \sin(\theta) \cos(\alpha)
\end{align}
where $\tau_x, \tau_y$ are the centroid.
$\rho_a,\rho_b$ are the axes lengths.
$\theta$ is the parameter belong to $[0, 2 \pi]$.
$\alpha$ is the rotation angle.
Let $\theta_i = 2 \pi i / 1800, i\in[1800]$, the generated 2D ellipse is given as
$^{\mathcal{F}_m} \Omega^e := \{ (x_i, y_i)  |  \theta_i \} \in \mathbb{R}^{1800 \times 2}$, and the $xy$-coordinates of $^{\mathcal{F}_m} \mathbf{V}$ is extracted as $^{\mathcal{F}_m} \Omega^v :=  \in \mathbb{R}^{n_v \times 2}$.

The 2D ellipse standard equation is constructed as:
\begin{align}
\label{eq19}
&
^{\mathcal{F}_m} f_{s}(x,y)
= {{{{\left( {\left( {x - {\tau_x}} \right)\cos \alpha  + \left( {y - {\tau_y}} \right)\sin \alpha } \right)}^2}}} / {{{ \rho_a ^2}}} \notag \\
&+ {{{{\left( {\left( {{\tau_x} - x} \right)\sin \alpha  + \left( {y - {\tau_y}} \right)\cos \alpha } \right)}^2}}} / {{{ \rho_b ^2}}} \ge 0
\end{align}

Whether $(x,y)$ is inside the ellipse can be judged by \eqref{eq19}, which is
used construct the subsequent constraint.
Let the perimeter of the bag's rim as $\omega$, and the cost function that satisfies the perimeter limitation is constructed as:
\begin{equation}
\label{eq20}
\mathcal{J}_1(\tau_x, \tau_y, \rho_a, \rho_b, \alpha) 
= 
\big\| 
2\pi \sqrt{ {  (\rho_a^2 + \rho_b^2 ) }/{2}} - \omega 
\big\|^2
\end{equation}

Further, three additional constraints are defined as:

{\textbf{Constraint $C_1$:}}
it regulates the covering of $^{\mathcal{F}_m } \Omega ^e$ to $^{\mathcal{F}_m } \Omega ^v$:
\begin{equation}
\label{eq23}
0 \le ^{\mathcal{F}_m} \hspace{-0.1cm} f_s 
(^{\mathcal{F}_m } v_{i,x}, ^{\mathcal{F}_m } v_{i,y}) \le \lambda_1,  \ \ \
i \in [1,\ldots, n_v]
\end{equation}
where $\lambda_1$ controls the enclosing degree.
The smaller $\lambda_1$ is, the center alignment between $^{\mathcal{F}_m} \Omega^e$ and $^{\mathcal{F}_m} \Omega^v$ more obvious.
The default value is $\lambda_1 = 0.87$.

{\textbf{Constraint $C_2$:}} it limits the Euclidean distance between the centers of $^{\mathcal{F}_m} \Omega^e$ and $^{\mathcal{F}_m} \Omega^v$:
\begin{equation}
\label{eq22}
0 \le
\big\| 
[\tau_x, \tau_y, 0] -  ^{\mathcal{F}_m} \bar{\mathbf{v}} 
\big\| \le \lambda_2
\end{equation}
where $\lambda_2$ specifies the proximity of the two centers, it has the similar control effect to $\lambda_1$.
The default value is $\lambda_2 = 0.003$.

{\textbf{Constraint $C_3$:}} 
it adjusts the parallelism of the respective principal axes of
$^{\mathcal{F}_m} \Omega^e$ and $^{\mathcal{F}_m} \Omega^v$,
denoted as $
\eta_{e} \in \mathbb{R}^2, \eta_{v} \in \mathbb{R}^2$ and calculated by PCA \cite{hasan2021review}.
Afterwards, the inner product is used to evaluate the parallelism:
\begin{align}
\label{eq26}
- \lambda_3 \le 
| {\rm{dot}}(  \eta_{e} ,  \eta_{v} ) |  - 1  
\le \lambda_3
\end{align}
where $\lambda_3$ controls the parallel degree.
As we only consider the parallelism, and ignore the direction (same/opposite), so we take absolute operation and subtract 1.
The default value is $\lambda_3 = 0.0001$.

The optimal values $(\tau_x^*,\tau_y^*, \rho_a^*,\rho_b^*, \alpha^*)$ can be obtained by minimizing $\mathcal{J}_1$, and considering three constraints $C_1, C_2, C_3$.
The nonlinear optimizer is adopted to obtain the optimal values:
$^{\mathcal{F}_m} \Omega ^{e,\ast} := 
\{ (x_i, y_i) |  \tau_x^*, \tau_{y}^{*}, \rho _{a}^{*}, \rho _{b}^{*}, \alpha^* \}$, then concatenate a zero vector horizontally to make 
$^{\mathcal{F}_m} \Omega ^{e,\ast}$ three-dimensional.

\textbf{Step 3:} \emph{Bagging SOI Generation.}

Similarly, adopting \eqref{eq16} to map $^{\mathcal{F}_m} \Omega ^{e,\ast}$ into $\mathcal{F}_w$ to obtain $^{\mathcal{F}_w} \Omega ^{e,\ast}$, and whose dimension should be consistent with that of $\mathbf{x}$.
Thus, farthest point sampling (FPS) \cite{yan2020pointasnl} extracts $n_x$ samples from $^{\mathcal{F}_w} \Omega ^{e,\ast}$ to obtain:
\begin{equation}
\label{eq31}
\mathbf{x}_{\dag} = {\rm{FPS}} (^{\mathcal{F}_w} \Omega ^{e,\ast}, \ n_x) 
\in \mathbb{R}^{n_x \times 3}
\end{equation}

\textbf{Step 4:} \emph{Goal SOI Generation.}

Note that $\mathbf{x}_{\dag}$ is coplanar with $\mathbf{V}$, and seen as an transient shape, i.e., $\mathbf{x}_{\dag}$ is at the bottom of $\mathbf{B}$.
Our goal is to generate a shape that surrounds the bottom part of $\mathbf{B}$, so we simply translate $\mathbf{x}_{\dag}$ along $\mathbf{a}$ by a safety threshold $\gamma$ to get $\mathbf{x}_{\ast}$, it yields
\begin{equation}
\label{eq46}
\mathbf{x}_{\ast} = \mathbf{x}_{\dag} + \gamma \cdot \mathbf{a}
\end{equation}

Finally, $\mathbf{x}_{\ast}$ is the goal SOI, covering the bottom part of $\mathbf{B}$.
Fig. \ref{fig5} visualizes the bagging/goal SOI and three constraints.

\begin{remark}
In this work, $\mathbf{a}$ should hold an acute angle with the positive direction of the $z$-axis, which can be adjusted by calculating $\rm{dot}(\mathbf{a}, [0,0,1])$.
If negative, we should reverse ${\mathbf{a}}$.
\end{remark}

\begin{figure}[htbp]
\centering
\includegraphics[width=0.97\columnwidth]{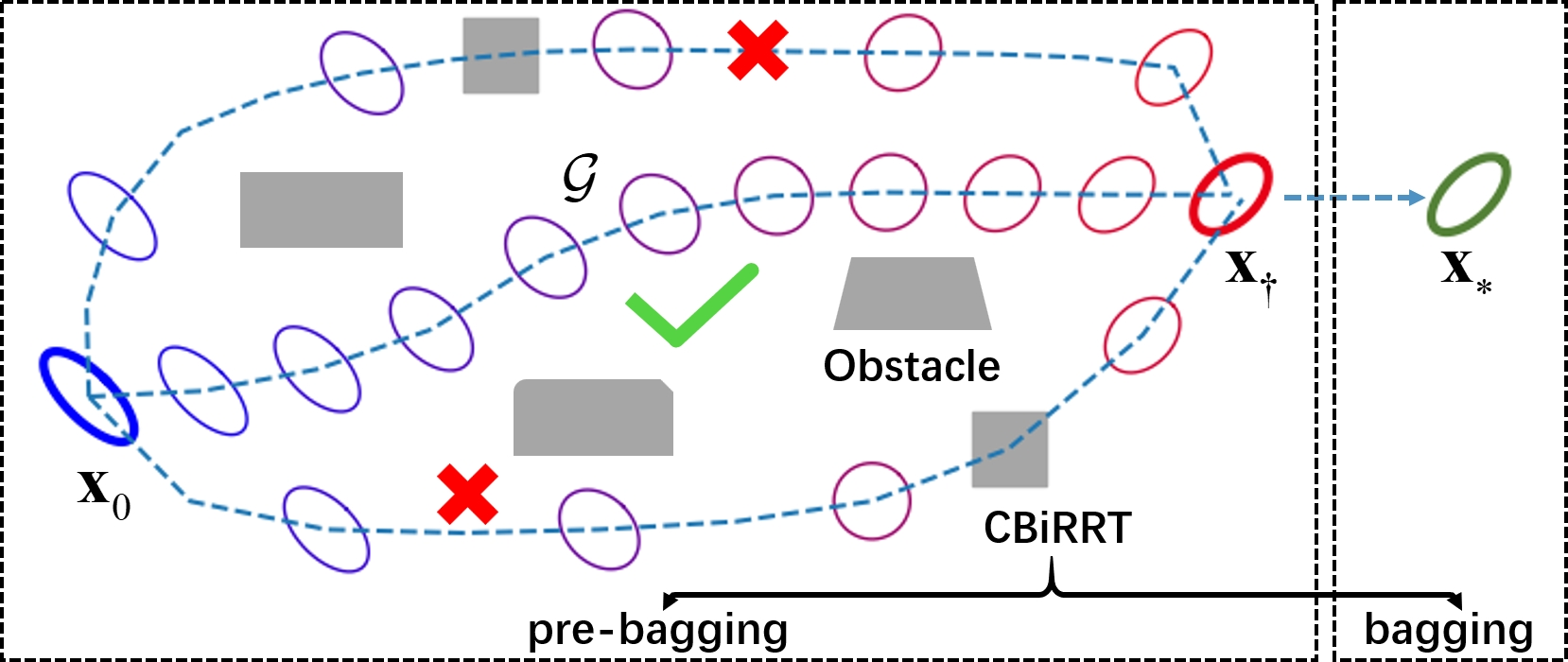}
\vspace{-0.2cm}
\caption{
Illustration of SOI planning.
$\mathbf{x}_0$ and $\mathbf{x}_{\dag}$ are the initial and bagging SOI.
$\mathbf{x}_{\ast}$ is goal SOI covering the bottom part of $\mathbf{B}$.
}
\vspace{-0.5cm}
\label{fig4}
\end{figure}

\subsection{SOI Planning}\label{section4c}

In this section, we introduce how to generate a collision-free deformation path $\mathcal{G}$ of the bag from the initial SOI $\mathbf{x}_0$ to $\mathbf{x}_{\ast}$, where $\mathbf{x}_{\ast}$ is the bagging SOI as the pre-enclosing configuration of $\mathbf{B}$.
Note that $\mathcal{G}$ includes two stages, $\mathbf{x}_0 \rightarrow \mathbf{x}_{\dag}$ and $\mathbf{x}_{\dag} \rightarrow \mathbf{x}_{\ast}$
Then, $\mathcal{G}$ serves as the desired trajectory of the subsequent controller to assist in completing the bagging task.
The SOI planning is conducted in the world frame $\mathcal{F}_w$.
The studied planning task can be seen as the shape planning of bag's SOI, as a global trajectory guiding the robot.

The 3D ellipse parametric equation is constructed as:
\begin{equation}
\label{eq34}
^{\mathcal{F}_w} f_{p} (\mathbf{c}, \beta_a, \beta_b, {\mathbf{u}}, {\mathbf{v}})
= 
\mathbf{c} + 
\beta_a \cos(\theta) {\mathbf{u}} + \beta_b \sin(\theta) {\mathbf{v}}
\end{equation}
where $\mathbf{c} = [c_x,c_y, c_z]$ is the centroid.
$\beta_a$ and $\beta_b$ determine the semi-major and semi-minor axes lengths, respectively.
$\theta$ is the parametric angle belong to $[0, 2 \pi]$.
${\mathbf{u}} \in \mathbb{R}^3, {\mathbf{v}}\in \mathbb{R}^3$ are the direction vectors.
Let $\theta_i = 2\pi i/2000, i\in[1,2000]$, and the 3-dimension ellipse  in $\mathcal{F}_w$ is constructed as:
\begin{equation}
\label{eq36}
\Upsilon_{e} :=
\{
\Upsilon_i = 
(x_i, y_i, z_i) | \theta_i \leftarrow \eqref{eq34}
\} 
\in \mathbb{R}^{2000 \times 3}
\end{equation}

The perimeter of $\Upsilon_{e} $ is numerically obtained as $\chi = \Sigma_{i=1}^{2001} \| \Upsilon_i - \Upsilon_{i-1} \|$ with $\Upsilon_{2001} = \Upsilon_{1}$ for the circle calculation.

\vspace{-0.2cm}
\begin{figure}[htbp]
\centering
\includegraphics[width=0.97\columnwidth]{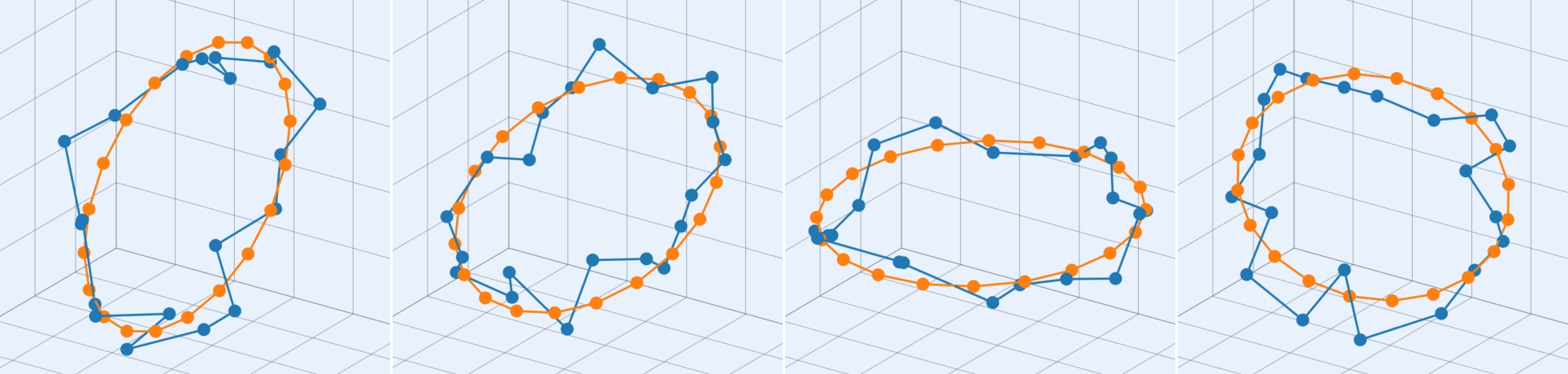}
\vspace{-0.4cm}
\caption{
Visualization of mapping a randomly sampled configuration in the raw space into the stable manifold.
The blue dot-line represents the randomly sampled configuration, and the yellow one is the stable configuration in the manifold through ProjectStableConfig.
}
\vspace{-0.3cm}
\label{fig3}
\end{figure}

\textbf{Projection of Stable Configuration Manifold}

The bag's raw configuration space $\mathbf{x}_t$ has a dimensionality of $3n_x$. 
However, its stable state is confined to a specific subspace known as a manifold within this larger space.
Therefore, it can enhance the planning credibility if the planning process is performed specifically on this manifold that contains the stable state of the bag.
However, it's challenging to obtain this manifold through random sampling in the raw space as the dimensions of the raw space significantly exceed those of the stable space.
To discover this constraint manifold within the current shape configuration, it would be suitable to employ the projection method, which allows for a more targeted exploration of the stable state \cite{berenson2009manipulation}.

By formulating a random sampling in the raw space as a local minimization problem of the energy, it can project it onto the stable manifold, with $\mathbf{x}_t$ as the initial value.
\begin{align}
\label{eq40}
\mathbf{x}_t^{\rm{stable}} 
&= \mathop{\arg\min}_{\mathbf{x}}
\ \mathcal{J}_2(\mathbf{x}) , \ \ \ 
{\rm{s.t.}} \ \ 
\mathbf{x}  = \mathbf{x}_t
\end{align}

A geometric index is used as the projection model, where the cost function $\mathcal{J}_2$ for a configuration $\mathbf{x}$ is presented as:
\begin{align}
\label{eq35}
\mathcal{J}_2
(
\mathbf{c}, \beta_a, \beta_b, {\mathbf{u}}, {\mathbf{v}}
) = 
|{\rm{CD}} (\Upsilon_{e}, \mathbf{x}_t)|^2
\end{align}
where $\rm{CD}(\cdot)$ is the Chamfer Distance, to evaluate the similarity between two unordered dataset with different dimensions.

{\textbf{Constraint $C_4$:}} it ensure the perimeter $\chi$ of $\Upsilon_{e}$ on the manifold be consistent with the bag rim's perimeter $\omega$.
\begin{equation}
\label{eq39}
1 - \lambda_4 \le
{\chi} \  / \ {\omega} \le 1 + \lambda_4
\end{equation}
where $\lambda_4$ controls the scale between $\chi$ and the ground-truth perimeter $\omega$.
The default value is set to $\lambda_4 = 0.001$.

{\textbf{Constraint $C_5$:}}
it ensures that the projection process completed without large offsets.
\begin{equation}
0\le \|  \mathbf{c} - \bar{ \mathbf{x}}_t  \| < \lambda_5
\end{equation}
where $\bar{\mathbf{x}}_t$ is the centroid of $\mathbf{x}_t$.
The default value is $\lambda_5 = 0.01$.

The optimal values of \eqref{eq34} can be obtained by minimizing $\mathcal{J}_2$  with \eqref{eq39}.
Afterwards, $\mathbf{x}_t^{\rm{stable}} = 
{^{\mathcal{F}_w} f_p }
(\mathbf{c}^{*},\beta_a^{*},\beta _{b}^{\ast},{\mathbf{u}}^{*},{{{\mathbf{v}} }^{\ast}})$.
Additional constraints can indeed be incorporated to cater to specific tasks.
The projection process from a raw configuration space to a neighboring stable manifold is denoted as $\mathbf{x}^{\rm{stable}} = \rm{ProjectStableConfig}(\mathbf{x}_t)$.
Four examples of ProjectStableConfig are shown in Fig. \ref{fig3}.

\textbf{Step 1: Pre-Bagging SOI Planning}

Our shape planning algorithm follows the same streamline as the Constrained Bi-directional Rapidly-Exploring Random Tree (CBiRRT) \cite{berenson2009manipulation}.
ProjectStableConfig ensures the validity of nodes, and CBiRRT contains two independent trees, growing from the initial configuration and the goal configuration, respectively. 
Both trees expand and explore the configuration space, gradually moving towards each other, until they eventually become connected to generate the final path. 
For our planning, the bag's state $\mathbf{x}_t$ is regarded as the tree's node.
The procedure of CBiRRT is introduced in \cite{yu2023coarse}, interesting readers could refer to it.

In planning, each random node $\mathcal{G}_{\rm{rand}}$ is projected to the stable manifold using ProjectStableConfig before the next-step planning.
If the bag is in collision, $\mathcal{G}_{\rm{rand}}$ is discarded and regenerated.
The Chamfer Distance is utilized to calculate the distance between two nodes, this point is different from \cite{yu2023coarse}.

Similiiar to \cite{berenson2009manipulation}, the constrained extension is denoted as $\mathcal{G}_{\rm{reached}}$ = {ConstrainedExtend}($\mathcal{G}_{\rm{from}}, \mathcal{G}_{\rm{to}})$.
This function aims to make progress from $\mathcal{G}_{\rm{from}}$ towards reaching $\mathcal{G}_{\rm{to}}$ while adhering to the constraints and limitations imposed by the planning problem.
During each step of the process, a new configuration for the bag is generated by interpolating from the last reached configuration $\mathbf{x}_{\rm{last}}$ to $\mathbf{x}_{\rm{to}}$, using a small step size. 
To ensure the overall shape of the bag is preserved and prevent excessive stretching, the displacement limitation for the relative deformation of the bag's rim is enforced. 
Afterwards, a stable configuration $\mathbf{x}_{\rm{new}}$ is obtained using {ProjectStableConfig} on the interpolated configuration.

After planning, the bidirectional path of CBiRRT is extracted as the deformation path $\mathcal{G}$, then refined for the subsequent smooth dual-arm manipulation.
The pre-bagging path is constructed as $\mathcal{G}_{\rm{pre\text{-}bagging}} := 
\{ \mathbf{g}_0 , \mathbf{g}_1, \ldots, \mathbf{g}_{\dag} \} $

\textbf{Step 2: Bagging SOI Planning}

This stage presents the deformation path from $\mathbf{x}_{\dag}$ to $\mathbf{x}_{\ast}$, adopting the same planning procedure as the \textbf{Step 1}.
The bagging path is constructed as $\mathcal{G}_{\rm{bagging}} := 
\{ \mathbf{g}_{\dag}, \ldots , \mathbf{g}_{\ast}\}$.

The final path $\mathcal{G}$ from $\mathbf{x}_0$ to $\mathbf{x}_{\ast}$ is constructed as:
\begin{equation}
\label{eq47}
\mathcal{G} 
:= 
\{ {\left[ \mathcal{G}_{\rm{pre \text{-} bagging}}  ,  \mathcal{G}_{\rm{bagging}} \right]  
| \underbrace {\mathbf{g}_0,\mathbf{g}_1, \ldots ,
\mathbf{g}_{\dag}}_{\rm{pre \text{-} bagging}},
\underbrace {\mathbf{g}_{\dag}, \ldots , \mathbf{g} _{\ast}  }_{\rm{bagging}}} \}
\end{equation}
\vspace{-0.2cm}

\vspace{-0.4cm}
\subsection{Motion Planning}
\label{section4d}
The process of the proposed bagging manipulation approach is:
(a) the vision system perceives $\mathbf{B}$, and the robot generates the bagging/goal SOI.
(b) an collision-free deformation path $\mathcal{G}$ is obtained using CBiRRT.
(c) the dual robot completes the bagging task along $\mathcal{G}$ in a constrained environment.
For providing a clear and intuitive visual effect, the robot adopts 3D translation and 3D rotation.
The end-effector's pose is denoted by $\mathbf{r} = [\mathbf{p}^{\rm{[left]}}, \mathbf{p}^{[\rm{right}]}] \in \mathbb{R}^{12}$.
We assume that the material properties of the bag and the robot movements remain relatively stable during the manipulation process.
The robot is able to execute  the given velocity commands accurately and without delay.
For the controller design, we formulate this manipulation process as the shape servoing \cite{qi2023adaptive}, i.e., tiny movements of the robot can produce tiny deformations of the bag.
Inspired by \cite{qi2021contour}, the local first-order kinematic model can be obtained as below:
\begin{equation}
\label{eq4}
\mathbf{y}_t = \mathbf{J}_t  \mathbf{u}_t, \ \
\mathbf{y}_t = \mathbf{x}_t - \mathbf{x}_{t-1}, \ \
\mathbf{u}_t = \mathbf{r}_t - \mathbf{r}_{t-1}
\end{equation}
where $\mathbf{J}_t$ is the deformation Jacobian matrix (DJM), which represents the kinematic relationship between $\mathbf{y}_t$ and $\mathbf{u}_t$.
We make the assumption that $\mathbf{J}_t$ maintains full column rank while performing the manipulation task, which is straightforward to fulfill in practical scenarios since the dimension of $\mathbf{x}$ is significantly greater than $\mathbf{u}$.
Since the bag has strong unknown nonlinearity, it's difficult to obtain accurate analytical expression of $\mathbf{J}_t$.
Therefore, the Broyden approach is used to computes local approximations of $\mathbf J_t$ in real time instead of identifying the full mechanical model.
\begin{equation}
\label{eq41}
\hat{\mathbf{J}}_{t} = 
\hat{\mathbf{J}}_{t-1} +
\varepsilon \cdot
({{ {{\mathbf{y}_t} - \hat{\mathbf{J}}_{t-1}{\mathbf{u}_t}}}}) \ / \ 
({{\mathbf{u}_t^\T {\mathbf{u}_t}}}) \cdot 
\mathbf{u}_t^\T
\in \mathbb{R}^{3n_x \times 12}
\end{equation}
where $\varepsilon \in (0,1]$ regulates the convergence speed.

Considering $\mathcal{G}$ contains a sequence of trajectories, thus we adopt a model predictive control (MPC) to drive the dual-arm manipulate along $\mathcal{G}$.
To simplify calculation burden, $\mathbf{J}_t$ is assumed to be estimated accurately, such that it satisfies $ \mathbf{y}_{t} = {\hat{\mathbf{J}}_t}{\mathbf{u}_t}$.
Two prediction vectors are defined as follows:
\begin{align}
\label{eq42}
\bar{\mathbf{x}}_t = \{ \mathbf{x}_{t + i|t}  \} \in {\mathbb{R}^{3 n_x h}}, \
\bar{\mathbf{u}}_t = \{ \mathbf{u}_{t+i-1|t}  \} \in {\mathbb{R}^{12h}}, 
i \in [1,h]
\end{align}
where $\bar{\mathbf{x}}_t$ and $\bar{\mathbf{u}}_t$ represent the predictions of $\mathbf{x}_t$ and $\mathbf{u}_t$ in the next $h$ periods, respectively.
$\mathbf{x}_{t+i|t}$ and $\mathbf{u}_{t+i|t}$ denote the $i$th predictions of $\mathbf{x}_t$ and $\mathbf{u}_t$ from the time instant $t$, where $\mathbf{x}_{t|t} = \mathbf{x}_t$, and $\mathbf{u}_{t|t} = \mathbf{u}_t$ must hold.
$\bar{\mathbf{x}}_t$ can be calculated from $\hat{\mathbf{J}}_t$ by noting that $\hat{\mathbf{\mathbf{J}}}_t\approx\hat{\mathbf{\mathbf{J}}}_{t+h}$ is satisfied during period $[t,t+h]$ (which is reasonable, given the slow manipulation of the bag).
In this way, $\bar{\mathbf{x}}_t$ are computed as the augmented format: 
\begin{align}
\label{eq43}
\bar{\mathbf{x}}_t
&= \mathbf{D} \mathbf{x}_t + \boldsymbol{\Theta} {\bar{\mathbf{u}}_t}  ,  \ \ \
\mathbf{D}  = \mathbf{I}_{h \times 1} \otimes \mathbf{E}_{3n_x}, \ \ \
\boldsymbol{\Theta} = \mathbf{L}_h \otimes \hat{\mathbf{J}}_t
\end{align}

The target $\bar{\mathbf{x}}_t^*$ is constructed as $\bar{\mathbf{x}}_t^* 
=[
\mathcal{G}_{t + 1},
\mathcal{G}_{t + 2}, \ldots,
\mathcal{G}_{t + h}
]$.
The optimization function of $\bar{\mathbf{u}}_t$ is formulated as:
\begin{align}
\label{eq37}
\mathcal{Q}\left( \bar{\mathbf{u}}_t \right) 
= {\left( {{\bar{\mathbf{x}}_t} - {\bar{\mathbf{x}}_t^*}} \right)^\T} \boldsymbol{\Lambda}_1 \left( {{\bar{\mathbf{x}}_t} - {\bar{\mathbf{x}}_t^*}} \right) + \bar{\mathbf{u}}_t^\T \boldsymbol{\Lambda}_2 {\bar{\mathbf{u}}_t}
\end{align}
where $\boldsymbol{\Lambda}_1$ and $\boldsymbol{\Lambda}_2$ are symmetric positive-definite matrices, regulating the convergence speed and the smoothness of $\bar{\mathbf{u}}_t$, respectively.
Finally, ${\mathbf{u}}_t$ is obtained by the receding horizon:
\begin{align}
\label{eq45}
\mathbf{u}_t = [\mathbf{E}_{12},\mathbf{0},\ldots,\mathbf{0}] \cdot \bar{\mathbf{u}}_t \in \mathbb{R}^{12}
\end{align}

\vspace{-0.5cm}
\begin{figure}[htbp]
\centering
\includegraphics[width=0.95\columnwidth]{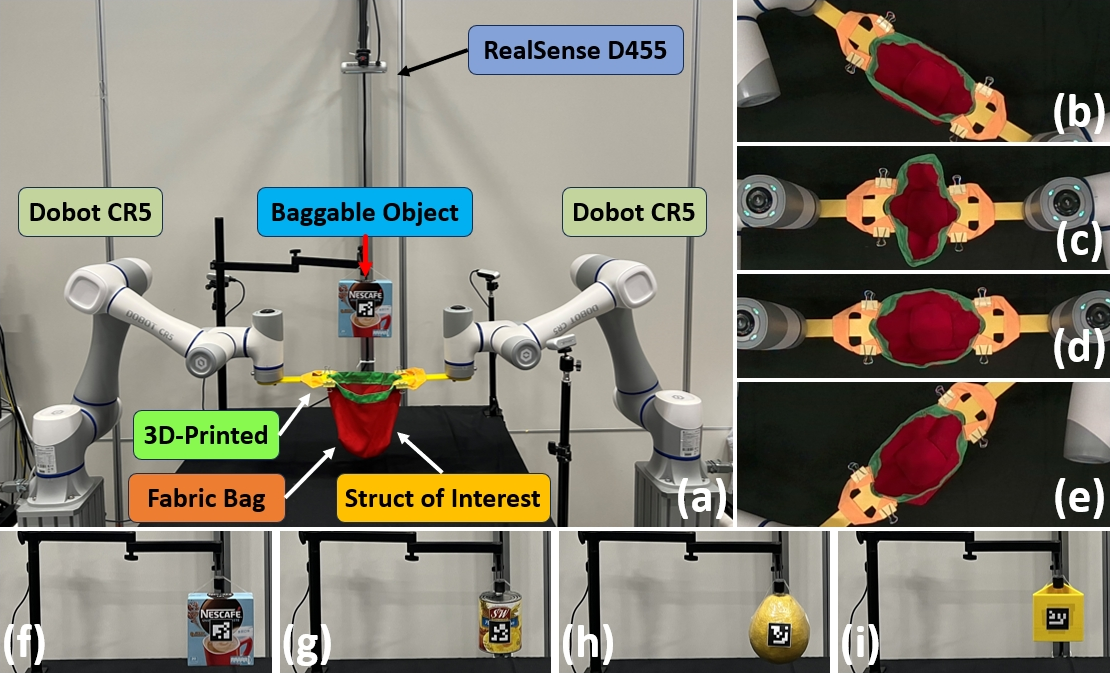}
\vspace{-0.2cm}
\caption{
\textbf{Experimental set-up.}
(a) Overview of the experimental set-up to validate our SOI-based control for the dual-arm bagging task. 
(b)-(e) Four samples of dual-arm manipulation bags.
(f) - (i) Four baggable objects, i.e., 
coffee box, canned pineapple, grapefruit, and 3D-printed triangular prism.
}
\label{fig6}
\end{figure}

\vspace{-0.3cm}
\section{Experiments}\label{section7}

\subsection{Experimental Setup}\label{section7a}
As shown in Fig. \ref{fig6}, we describe The experimental setup used to validate the proposed \textbf{SOI-based} control of dual-arm bagging task, including four baggable objects $\mathbf{B}$.
A D455 camera is in the eye-to-hand configuration, and used to observe the manipulation process from a top-down perspective with the resolution 640x480.
Visual perception is processed with OpenCV on a Linux-based PC, and the point cloud $\mathcal{P}_t$ are obtained through RealSense libraries.
Dual-CR5 robots are equipped with 3D-printer holders to grasp the both ends of the bag with zip ties in advance, and assume that no drops occur during manipulation.
The custom bag with a green rim is adopted for ease of perception.
The velocity command $\mathbf{u}_t$ has a hard saturation to meet the assumption in Sec. \ref{section3} to ensure the estimation validity of $\mathbf{J}_t$.
The motion control algorithm is implemented on ROS, which runs with a servo-control loop of around 11 Hz.

We use professional 3D scanners (Model: CR-Scan Ferret Pro) to obtain the vertex points $\mathbf{V}$ of each $\mathbf{B}$.
Meanwhile, the ArUco markers are attached to $\mathbf{B}$ to ensure that the robot can determine the type through the camera before manipulation, then call the corresponding configuration of $\mathbf{V}$.

\subsection{Evaluation of GMM-based State Estimation}
In this section, we verify the GMM-based state estimator introduced in \eqref{eq10}, it aims to extract clear state $\mathcal{Q}_t$ from the raw dense and noisy point cloud $\mathcal{P}_t$.
As the used bag has an obvious rim, so $\mathcal{P}_t$ can be obtained simply.
EM algorithm \cite{tang2022track} is used to solve $\mathcal{O}(\mathbf{x}_t^n) $, we can get the concise $\mathcal{Q}_t$.

\begin{figure}[htbp]
\centering
\includegraphics[width=0.97\columnwidth]{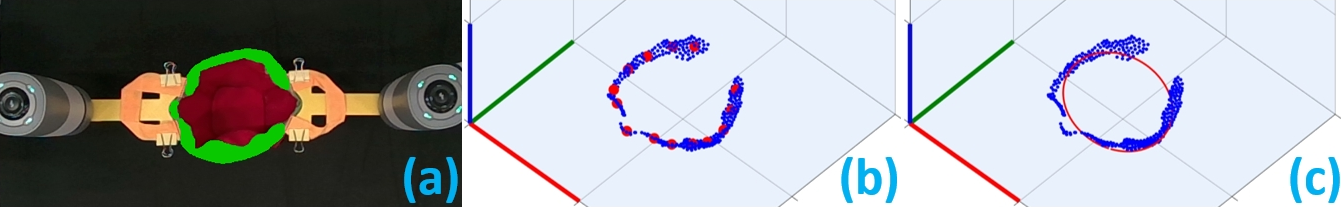}
\vspace{-0.4cm}
\caption{
Extraction results of the GMM-based state estimator.
(a) the bag's SOI.
(b) the blue dots are $\mathcal{P}_t$, and the red ones are $\mathcal{Q}_t$ estimated by GMM.
(c) the smooth state by refining $\mathcal{Q}_t$ using ProjectStableConfig.
}
\vspace{-0.6cm}
\label{fig7}
\end{figure}

Fig. \ref{fig7} shows the extraction effect of the GMM-based state estimator, and the bag's rim are marked by green, as shown in Fig. \ref{fig7}a.
The results in Fig. \ref{fig7}b show that the GMM-based state estimator can propose a relatively completely $\mathcal{Q}_t$, and $\mathcal{Q}_t$ is equidistantly distributed, this echoes the uniform distribution assumption \eqref{eq48}.
Furthermore, Fig. \ref{fig7}c is added to evaluate ProjectStableConfig in \eqref{eq40}.
The results show that ProjectStableConfig can find a stable manifold under the current shape configuration $\mathcal{Q}_t$, presented as the red curve in Fig. \ref{fig7}c distributed along $\mathcal{Q}_t$.
This proves the effectiveness of ProjectStableConfig and can find a stable manifold projection, which is helpful for subsequent planning and control.

\vspace{-0.2cm}

\begin{figure*}[htbp]
\centering
\includegraphics[width=0.92\textwidth]{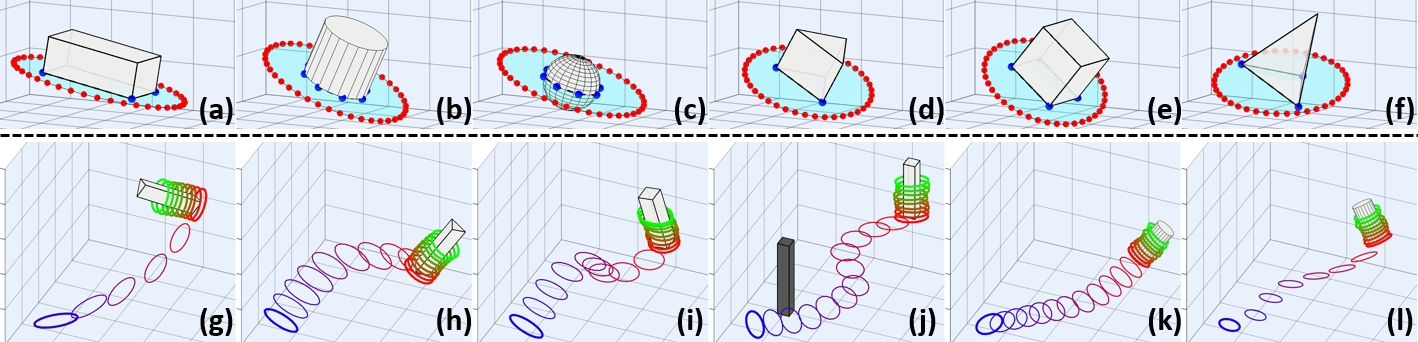}
\vspace{-0.3cm}
\caption{
(a) - (f): Generations of bagging SOI $\mathbf{x}_{\dag}$.
(g)-(l):  Deformation path $\mathcal{G}$.
}
\label{fig15}
\end{figure*}

\subsection{Evaluation of bagging SOI Generation}
In this section, we evaluate the bagging SOI generation presented in Sec. \ref{section4b}, to generate a pre-enclosed shape $\mathbf{x}_{\dag}$ to cover the bottom of $\mathbf{B}$.
Six types of baggable objects are adopted with the known $\mathbf{V}$.
The parameters are set to $\omega=0.68, \lambda_1 = 0.85, \lambda_2 = 0.005, \lambda_3 = 0.001$.

Fig. \ref{fig15}a - Fig. \ref{fig15}f shows the results of the bagging SOI $\mathbf{x}_{\dag}$.
The blue points represent $\mathbf{V}$, and the red one is $\mathbf{x}_{\dag}$ through \eqref{eq31}.
The results show that $\mathbf{x}_{\dag}$ satisfies the perimeter $\omega$, and can surround $\mathbf{V}$ to the greatest extent along the principal axis of $\mathbf{V}$.
And $\mathbf{x}_{\dag}$ is generated evenly distributed around $\mathbf{V}$, which verifies the regulation of $C_1$.
As $\mathbf{x}_{\dag}$ is generated by a parametric equation, $\mathbf{x}_{\dag}$ is continuous, which is helpful for subsequent SOI planning.
Moreover, in the experiment, we found that by adjusting $\lambda_1$ of constraint $C_1$ \eqref{eq23} and $\lambda_3$ of constraint $C_3$ \eqref{eq26}, it can efficiently regulate $\mathbf{x}_{\dag}$ to adapt to various $\mathbf{B}$, so that $\mathbf{x}_{\dag}$ can best meet different task requirements.
The average values of $C_1$, $C_2$, and $C_3$ are 0.731, 0.004, and $0.0001$, respectively.





\vspace{-0.2cm}
\subsection{Evaluation of SOI Planning}
In this section, we evaluate the SOI planning presented in Sec. \ref{section4c}, which aims to generate a collision-free deformation path $\mathcal{G}$ from the initial SOI $\mathbf{x}_0$ to the goal SOI $\mathbf{x}_{*}$ via the bagging SOI $\mathbf{x}_{\dag}$.
The parameter is $\lambda_4 = 0.002, \lambda_5=0.02$.

Fig. \ref{fig15}g - Fig. \ref{fig15}l give six planning results $\mathcal{G}$ of different configurations of $(\mathbf{x}_0, \mathbf{x}_{\dag}, \mathbf{x}_{\ast})$.
The gradient curves from blue to red is $\mathcal{G}_{\rm{pre \text{-} bagging}}$, and that from red to green is $\mathcal{G}_{\rm{bagging}}$.
As ProjectStableConfig is used, thus each node in $\mathcal{G}$ is smooth, and satisfies the physical perimeter constraint $\omega$. 
Since the distance from $\mathbf{x}_{\dag}$ to $\mathbf{x}_{\ast}$ is short and the distance to $\mathbf{B}$ is close, $\mathcal{G}_{\rm{bagging}}$ shows a certain degree of fluctuation.
Fig. \ref{fig15}j shows that CBiRRT can generate an effective deformation path $\mathcal{G}$ even when there are obstacles.
The black cuboid is the used-defined obstacle.
The planning results show that continuous $\mathcal{G}$ can be obtained using the CBiRRT, and \eqref{eq39} guarantees the perimeter limitation of each node in $\mathcal{G}$.
This proves the rationality of the optimization manner \eqref{eq35}, and various constraints can be added to improve the planning accuracy and  meet the  task requirements.

Note that the proposed SOI planning is a top-level planning framework, the kind where the initial SOI $\mathbf{x}_0$, 
bagging SOI $\mathbf{x}_{\dag}$, and the goal $\mathbf{x}_{\ast}$ are given, the two-stage deformation trajectories are planned, i.e., $\mathcal{G}_{\rm{pre \text{-} bagging}}$ and $\mathcal{G}_{\rm{bagging}}$.
It's just that in this article, $\mathbf{x}_{\ast}$ is done by simply translating $\mathbf{x}_{\dag}$, but $\mathbf{x}_{\ast}$ may have more complex format actually.

\begin{figure*}[htbp]
\centering
\includegraphics[width=0.95\textwidth]{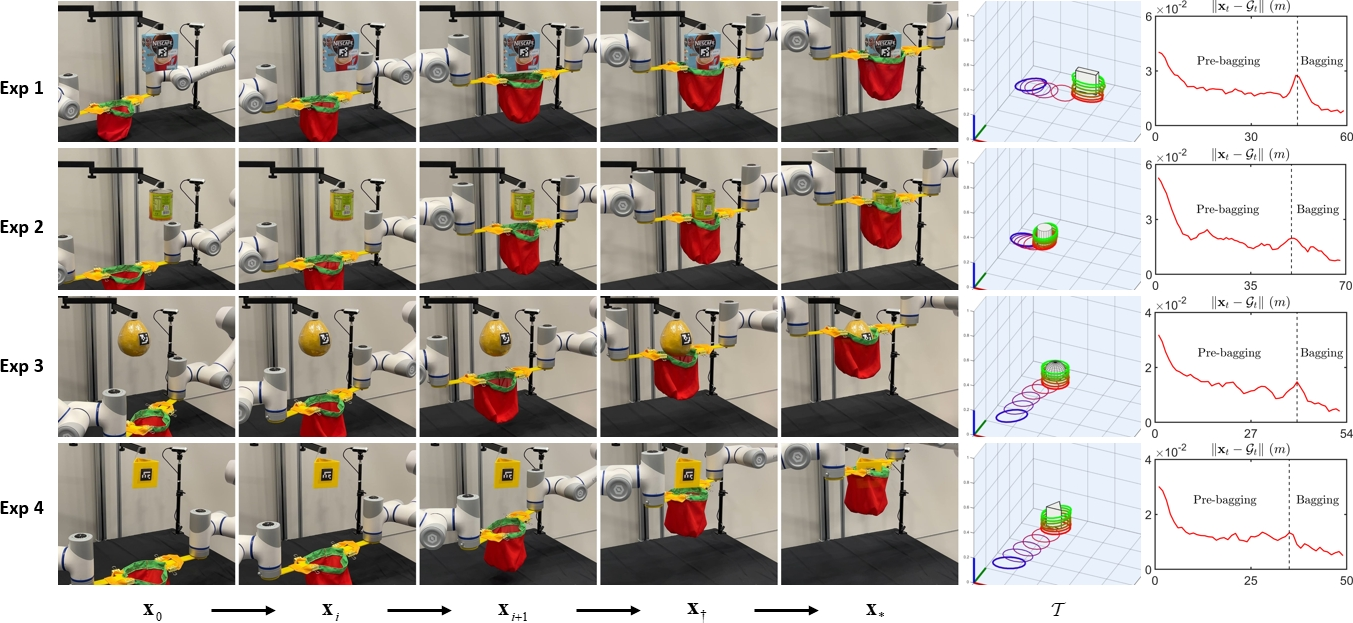}
\vspace{-0.4cm}
\caption{
Manipulation trajectories of the bag's SOI in the bagging task.
The last column gives the deformation error of each step of MPC.
}
\label{fig14}
\vspace{-0.4cm}
\end{figure*}

\subsection{Dual-arm Bagging Manipulation}
The dual-arm bagging experiments are conducted to evaluate the proposed SOI-based bagging manipulation approach.
The used baggable objects contain four types, i.e., coffee box, canned pineapple, grapefruit, and 3D-printed triangular prism, for Exp 1 to Exp 4, respectively.
The fundamental process is that the dual-CR5 manipulates the bag to first deform along $\mathcal{G}_{\rm{pre \text{-} bagging}}$ to $\mathbf{x}_{\dag}$, then deform along $\mathcal{G}_{\rm{bagging}}$ to $\mathbf{x}_{\ast}$, and finally complete the bagging task.
For analyzing the bagging approach,
we compare two planning algorithms (FFG-RRT \cite{roussel2014motion}, TS-RRT \cite{suh2011tangent})
and two manipulation algorithms (IBVS \cite{ren2020image}, SSVS \cite{hao2011universal}) respectively in Sec. \ref{section4c} and Sec. \ref{section4d}.

Fig. \ref{fig14} shows the bagging results of four experiments, with each row representing a type of $\mathbf{B}$.
The first five columns of each row represent the deformation process, the sixth column represents $\mathcal{G}$, and the last column represents the deformation error $\| \mathbf{x}_t - \mathcal{G}_t\|$ of each step of MPC.
In order to quantitatively compare performance, three indicators are introduced, i.e., planning success rate, planning time, and manipulation success rate, corresponding to different planning algorithms and control algorithms respectively.
Table \ref{table2} gives the detailed comparative analysis outcomes.

Planning success rate shows that CBiRRT outperforms the other counterpart, with the acceptable computation time, while FFG-RRT is the fastest.
This is because FFG-RRT directly explores forward and rushes to the desired configuration at the fastest speed, while CBiRRT conducts two-way exploration based on stability, this results in CBiRRT have more exploration steps.
From the manipulation success rate, we know that the MPC used in this article has the highest value, while the other two control approaches are slightly worse.
This is because the desired command of the traditional shape servoing is stationary, while that of our bagging task is actually a sequence of deformation trajectories.
This point is very consistent with the MPC processing manner, and can ensure the stability of tracking in the future prediction time domain.
The manipulation results prove the effectiveness of MPC in such robot manipulation tasks.

Besides, $\mathbf{x}_{\dag}$ is equivalent to an intermediate buffer shape, thus dividing the entire bagging task into two subtasks, namely \textbf{pre\text{-}bagging} and \textbf{bagging}, thereby improving the success rate of manipulation.

\begin{table*}[htbp]
\centering
\caption{Performance of Different Sensorimotor Models on Different Tasks for Motor-robot Experiments}
\label{table2} 
\scalebox{0.72}{   
\begin{tabular}{lcccccccccccccc} 
\toprule
\multirow{2}{*}{ Method } 
& \multicolumn{3}{c}{Coffee box (Exp 1)} 
& \multicolumn{3}{c}{Canned pineapple (Exp 2)}  
& \multicolumn{3}{c}{Grapefruit (Exp 3)} 
& \multicolumn{3}{c}{Triangular prism (Exp 4)} 
 \\
\cmidrule(lr){2 - 4} 
\cmidrule(lr){5 - 7}
\cmidrule(lr){8 - 10}
\cmidrule(lr){11 - 13}

& \makecell{Planning \\ success rate}
& \makecell{Planning \\ time (s)}
& \makecell{Manipulation \\success rate}

& \makecell{Planning \\ success rate}
& \makecell{Planning \\ time (s)}
& \makecell{Manipulation \\success rate}

& \makecell{Planning \\ success rate}
& \makecell{Planning \\ time (s)}
& \makecell{Manipulation \\success rate}

& \makecell{Planning \\ success rate}
& \makecell{Planning \\ time (s)}
& \makecell{Manipulation \\success rate}
\\
\midrule

FFG-RRT \cite{roussel2014motion}
& 6/10
& $3.87 \pm 1.97$
& 8/8

& 8/10
& 2.37 $\pm$ 0.87 
& 8/8

& 7/10
& 3.89 $\pm$ 1.18
& 8/8

& 6/10
& 3.58 $\pm$ 1.11
& 8/8
\\

TS-RRT \cite{suh2011tangent}
& 7/10
& 6.32 $\pm$ 1.08 
& 8/8

& 8/10
& 5.58 $\pm$  1.13
& 8/8

& 9/10
& 6.85 $\pm$ 0.56 
& 8/8

& 7/10
& 7.32 $\pm$ 1.34 
& 8/8
\\

IBVS \cite{ren2020image}
& -
& -
& 4/8

& -
& - 
& 7/8

& -
& - 
& 5/8

& -
& - 
& 6/8
\\

SSVS \cite{hao2011universal}
& -
& - 
& 5/8

& -
& - 
& 7/8

& -
& - 
& 6/8

& -
& - 
& 7/8
\\

\textbf{Ours}
& 9/10
& 5.13 $\pm$ 1.26 
& 8/8

& 10/10
& 4.21 $\pm$ 0.98 
& 8/8

& 10/10
& 4.98 $\pm$ 1.93 
& 8/8

& 9/10
& 5.32 $\pm$1.56 
& 8/8

\\
\bottomrule
\end{tabular}
}
\vspace{-0.4cm}
\end{table*}

\vspace{-0.2cm}
\section{Conclusion}
Our study introduced a dual-arm robotic system for automating bagging tasks, employing a novel constraint-aware SOI planning approach for manipulating 3D deformable objects. The system's innovation lies in its targeted SOI state estimation, which simplifies the control of the bag's opening rim, enhancing task efficiency. Key contributions include a flexible, adaptive vision-based control system and a comprehensive framework demonstrating the system's adaptability to environmental constraints. This research not only advances DOM in handling complex tasks but also has potential implications for enhancing robotic assistance in everyday activities. Future efforts will aim to improve system adaptability and extend its application to further realize the benefits of robotic automation in diverse real-world settings.




\bibliography{biblio-1}
\bibliographystyle{IEEEtran}

\end{document}